%% file: main.tex
\documentclass[twoside]{article}

\usepackage{aistats2019}
\usepackage{dsfont}
\usepackage{times}  %Required
\usepackage{helvet}  %Required
\usepackage{courier}  %Required
\usepackage{url}  %Required
\usepackage{graphicx}  %Required
\usepackage{color}

\usepackage[utf8]{inputenc} % allow utf-8 input
\usepackage[T1]{fontenc}    % use 8-bit T1 fonts
\usepackage{hyperref}       % hyperlinks
\usepackage{url}            % simple URL typesetting
\usepackage{booktabs}       % professional-quality tables
\usepackage{amsfonts}       % blackboard math symbols
\usepackage{nicefrac}       % compact symbols for 1/2, etc.
\usepackage{microtype}      % microtypography
\usepackage{tablefootnote}
\usepackage{threeparttable}
\usepackage{adjustbox}

\usepackage{subcaption}
\usepackage[colorinlistoftodos]{todonotes}
\usepackage{verbatim} % for block comment
\usepackage{amsmath}
\usepackage{grffile}
\usepackage{mathtools}
\usepackage{graphicx}
\usepackage{paralist}
\usepackage{bm} % bold math symbol
\usepackage{multirow}
\usepackage{makecell}
\usepackage[english]{babel}
\usepackage{bbm}
\usepackage{comment}
\usepackage{etoolbox}
\usepackage[noend]{algpseudocode} %%% Zhao add this
\usepackage{footnote}
\usepackage{enumitem}
\usepackage{xspace}
\usepackage{amssymb}

\usepackage{multirow}

\newcommand{\proven}{{\small \textsf{PROVEN}}\xspace}

\DeclareMathOperator*{\argmin}{argmin} 
\newcommand{\Conv}{%
  \mathop{\scalebox{1.5}{\raisebox{-0.2ex}{$\circledast$}}
  }
}

\begin{document}



\twocolumn[

\aistatstitle{PROVEN: Certifying Robustness of Neural Networks with a Probabilistic Approach}

\aistatsauthor{Tsui-Wei Weng\textsuperscript{1}, Pin-Yu Chen\textsuperscript{2}, Lam M. Nguyen\textsuperscript{2}, Mark S. Squillante\textsuperscript{2}, Ivan Oseledets\textsuperscript{3}, Luca Daniel\textsuperscript{1} \\
\textsuperscript{1}{MIT}, \textsuperscript{2}{IBM Research}, \textsuperscript{3}{Skoltech} }

\aistatsaddress{} 
]

\begin{abstract}
With deep neural networks providing state-of-the-art machine learning models for numerous machine learning tasks, quantifying the robustness of these models has become an important area of research. However, most of the research literature merely focuses on the \textit{worst-case} setting where the input of the neural network is perturbed with noises that are constrained within an $\ell_p$ ball; and several algorithms have been proposed to compute certified lower bounds of minimum adversarial distortion based on such worst-case analysis. In this paper, we address these limitations and extend the approach to a \textit{probabilistic} setting where the additive noises can follow a given distributional characterization. 
We propose a novel probabilistic framework \proven to \textbf{PRO}babilistically \textbf{VE}rify \textbf{N}eural networks with statistical guarantees -- i.e., \proven certifies the probability that the classifier's top-1 prediction cannot be altered under any constrained $\ell_p$ norm perturbation to a given input. Importantly, we show that it is possible to derive closed-form probabilistic certificates based on current state-of-the-art neural network robustness verification frameworks. Hence, the probabilistic certificates provided by \proven come naturally and with almost no overhead when obtaining the worst-case certified lower bounds from existing methods such as Fast-Lin, CROWN and CNN-Cert. Experiments on small and large MNIST and CIFAR neural network models demonstrate our probabilistic approach can achieve up to around $75\%$ improvement in the robustness certification with at least a $99.99\%$ confidence compared with the worst-case robustness certificate delivered by CROWN.  
\end{abstract}

%we show it is possible to derive closed-form probabilistic certificates in terms of neural network model parameters and input noises distribution.

\input{1_intro.tex}
\input{2_theory.tex}

\input{3_simu.tex}

\input{4_conclusion.tex}

% \clearpage

\clearpage

\clearpage
% \newpage
\bibliographystyle{IEEEtran}
\bibliography{reference}

% no need appendix
% \input{appendix.tex}

\end{document}

%% file: 1_intro.tex
\section{Introduction}
\label{sec:intro}
Despite the recent advances and successes of deep neural networks in many machine learning tasks, it has been shown that adversarial examples exist and can be easily crafted, spanning from image classification \cite{szegedy2013intriguing} to speech recognition \cite{cisse2017houdini} to malware detection \cite{wang2017adversary} and sparse regression \cite{chen2018ordered}, just to name a few. Although deep neural networks have achieved unprecedented performance in these applications, their lack of robustness against adversarial perturbations \cite{goodfellow2014explaining,biggio2017wild} has raised serious concerns and has drawn a great deal of attention by the machine learning communities, as many safety-critical tasks cannot afford the potential risks incurred by adversarial examples. 

While there is a growing interest in crafting adversarial examples with stronger attacks under various settings (e.g., white-box/grey-box/black-box attacks) and in developing effective defense strategies against adversarial attacks, the topic of assessing and verifying robustness properties of neural networks is equally important and challenging. Given a well-trained neural network model, we are interested in measuring its robustness on an arbitrary natural example $\mathbf x_0$ by examining if the neighborhood of $\mathbf x_0$ has the same prediction results; this serves as a robustness proxy for evaluating the ease with which one can turn $\mathbf x_0$ into adversarial examples via adversarial manipulations.
Conventionally, the concept of neighborhood is characterized by an $\ell_p$ ball centered at $\mathbf x_0$ with radius $\epsilon$ for any $p \geq 1$, where larger $\epsilon$ indicates greater robustness. Ideally, for robustness evaluation, we would like to find the smallest adversarial distortion imposed on $\mathbf x_0 $ that will change the model prediction, which is known as the \textit{minimum adversarial distortion}. Unfortunately, it has been shown that computing the minimum adversarial distortion on neural networks with ReLU activations is an NP-complete problem \cite{katz2017reluplex,sinha2017certifiable}, and hence formal verification methods such as Reluplex \cite{katz2017reluplex} and Planet \cite{ehlers2017formal} are computationally demanding and cannot scale to large realistic networks. 

% ============= Table of Notation ========

\begin{table*}[t]
\centering
\caption{Table of Notation}
\begin{tabular}{ll|ll}
\hline
\textbf{Notation} & \textbf{Definition} & \textbf{Notation} & \textbf{Definition} \\
\hline
$K$ & number of output classes &
CDF & cumulative distribution function \\
$f: \mathbb{R}^{n_0} \to \mathbb{R}^{K}$ & neural network classifier &
pdf & probability density function \\
$\mathbf{x_0} \in \mathbb{R}^{n_0}$ & original input vector &
$F_X$ & CDF of a random variable $X$ \\
$c = \argmin_i f_i(\mathbf{x_0})$ & predicted class of input $\mathbf{x_0}$ &
$f_X$ & pdf of a random variable $X$ \\
$g_t(\mathbf{x}) = f_c(\mathbf{x}) - f_t(\mathbf{x})$ & margin function at $\mathbf{x}$ for class $t$ &
$\mathbb{P}[g_t(X) > a]$ & probability that $g_t(X)$ is greater than $a$  \\
$g_t^L(\mathbf{x}): \mathbb{R}^{n_0} \to \mathbb{R}$ & linear lower bound of $g_t(\mathbf{x})$ &
$\gamma_L$ & theoretical lower bound of $\mathbb{P}[g_t(X) > a]$ \\
$g_t^U(\mathbf{x}): \mathbb{R}^{n_0} \to \mathbb{R}$ & linear upper bound of $g_t(\mathbf{x})$ &
$\gamma_U$ & theoretical upper bound of $\mathbb{P}[g_t(X) > a]$ \\
\hline 
\end{tabular}
\label{table_notation}
\end{table*}

% ========================================

As an alternative to minimum adversarial distortion, the concept of solving for a (non-trivial) \textit{lower bound} on minimum distortion as a certified robustness metric has been recently proposed in~\cite{hein2017formal,weng2018evaluating,kolter2017provable,raghunathan2018certified,weng2018towards,zhang2018efficient,Gehr2018AI2,Boopathy2019cnncert} and appears to be a promising approach. Theoretical lower bounds have been derived in terms of local Lipschitz constants for continuously-differentiable classifiers~\cite{hein2017formal} and neural networks with ReLU activations~\cite{weng2018towards}. In addition, there have been some recent works on developing algorithms that are able to deliver \textit{certified} lower bounds for fully-connected networks with ReLU activations~\cite{kolter2017provable,weng2018towards,Gehr2018AI2} and general activations~\cite{zhang2018efficient}, and for general convolutional neural networks with commonly-used convolutional layers, pooling layers and residual blocks~\cite{Boopathy2019cnncert}. Here the term \textit{certified} means that numerical values generated by these approaches are indeed deterministic lower bounds. In other words, all such approaches consider the setting where an input example can be perturbed by any perturbation bounded in an $\ell_p$ ball, and thus their analyses all belong to the category of \textit{worst-case} analysis.

On the other hand, additional alternative questions of interest include:
\begin{itemize}
    \item[(a)] What are the corresponding guarantees under the situation where the input data point is perturbed with some random noises? 
    \item[(b)] Can we provide confidence levels on the possibility that a given model will never be fooled under this probabilistic setting? 
\end{itemize}
One way to address questions such as (a) and (b) 
is to relate them to the sensitivity of a target model $f$ when the noise in the input is known to follow a given distribution. With prior knowledge on the noise distribution, this approach is expected to provide a more informative robustness certification in comparison to the prevailing worst-case analysis. More importantly, this statistical viewpoint of robustness indeed goes beyond the worst-case analysis considered in the adversarial attack setting. The probabilistic robustness certification is readily applicable to understanding the sensitivity and reliability of a target model subject to additive random noises under mild assumptions. For example, such random noises can be caused by data quantization, input preprocessing, or environmental background noises.

Unfortunately, to date there have been relatively little research efforts along these lines. Existing works \cite{franceschi2018robustness,fawzi2018adversarial} require some unverifiable or unrealistic assumptions on the classifier models and decision boundaries, rendering their results less useful in practice, especially for neural network models. This is indeed at the core of the motivation for our work -- we seek to develop a probabilistic framework that can address questions such as (a) and (b), without imposing unverifiable assumptions on the models or decision boundaries. 

In summary, we propose in this work a novel probabilistic framework \proven to \textbf{PRO}babilistically \textbf{VE}rify \textbf{N}eural network robustness. We show that it is possible to extend the conventional worst-case setting to a probabilistic setting based on existing worst-case certification frameworks with very little computational overhead, meaning that the probabilistic certificate comes naturally with nearly no additional overhead beyond worst-case robustness computations by methods such as Fast-Lin~\cite{weng2018towards}, CROWN~\cite{zhang2018efficient} and CNN-Cert~\cite{Boopathy2019cnncert}. 

%e extend the conventional worst-case setting to a probabilistic setting by only assuming the distribution of the input perturbations is zero-mean Gaussian or independent bounded random noise, and then derive theoretical bounds on the probability that a classifier's top-1 prediction cannot be altered under a constrained $\ell_p$ norm attack.
%this is a repetition from few lines above
%Notably, our framework does not rely on unverifiable assumptions imposed in the previous works, such as locally-approximately-flat decision boundaries \cite{franceschi2018robustness} or Gaussian distributed latent input vectors \cite{fawzi2018adversarial}.  

\paragraph{Contributions.} We highlight the contributions of this paper as follows.
\begin{itemize}
    \item A probabilistic framework \proven is proposed for certifying the robustness of neural networks under $\ell_p$ norm-ball bounded threat models, when the input noise follows a given distributional characterization (zero-mean Gaussian or independent bounded random noises). The established theoretical results are based on an $\ell_\infty$ constraint on the perturbation, but can be easily extended to other norms such as $\ell_1$ and $\ell_2$.
    
    \item Experimental results on large neural networks trained on MNIST and CIFAR datasets show that the robustness certification metric (i.e., the certified lower bound) can be greatly improved under the proposed probabilistic framework in comparison with the corresponding worst-case analysis results, even when the statistical risk is small. For example, with a confidence level of $99.99\%$, which means the robustness metric is almost 100$\%$ guaranteed to be certified, the improvement provided by our probabilistic framework over the worst-case analysis can be as high as $78.9\%$ for small networks and $32.8\%$ for large networks. 
    
    \item In addition to the noticeable improvement in the robustness metric, our probabilistic framework is a general tool that can be readily applied to neural networks with different activation functions, including tanh, sigmoid and arctan, as will be demonstrated in our experiments. Moreover, our proposed method is as computationally efficient as the worst-case analysis, since our probabilistic certificate has a closed-form and its parameters are by-products of worst-case certification frameworks (e.g., Fast-Lin~\cite{weng2018towards}, CROWN~\cite{zhang2018efficient}, and CNN-Cert~\cite{Boopathy2019cnncert}).
\end{itemize}

\section{Background and related works}
Given an input data example under a specified threat model, typically $\ell_p$ norm-ball bounded perturbation attacks, the goal of formal verification for adversarial robustness aims to certify a perturbation level $\epsilon$ such that the top-1 prediction will not be altered by any means. In other words, formal verification guarantees that no attack under the threat model can alter the top-1 prediction of the model if its attack perturbation is smaller than $\epsilon$. However, certifying the largest possible $\epsilon$, which is equivalent to finding the minimum perturbation required for a successful adversarial attack, has been shown to be an NP-complete problem \cite{katz2017reluplex} and thus it is computationally infeasible for large realistic networks. Alternatively, recent studies have shown that solving for a lower bound of the minimum perturbation for formal verification can be made more scalable and computationally efficient \cite{kolter2017provable,weng2018towards,Gehr2018AI2,dvijotham2018dual}. Some analytical lower bounds, based solely on model weights, have been derived \cite{szegedy2013intriguing,peck2017lower,hein2017formal,raghunathan2018certified} but they can be loose, even becoming trivial lower bounds (close to 0), or they only apply to 1 or 2 hidden layers. It is worth noting that current robustness verification approaches mainly focus on a ``worst-case'' analysis, whereas our  approach takes a probabilistic viewpoint for robustness certification. As will be evident in the following sections, our probabilistic framework approach \proven is able to certify a significantly larger $\epsilon$ value than the corresponding worst-case analysis result with $99.99\%$ certification guarantees. This indicates that, while the conventional worst-case robustness certification framework may be too conservative when there is some prior knowledge about the input perturbations (e.g., its distribution), our probabilistic framework will be more applicable in this situation.

In fact, deep neural networks are not only vulnerable to crafted adversarial noises but also to random noises: Bibi et al.~\cite{bibi2018analytic} show that they can fool LeNet and AlexNet with additive Gaussian noises and Fawzi et al.~\cite{fawzi2016robustness} show random perturbations can indeed fool VGG networks; Hosseini et al.~\cite{hosseini2017google} show they fool the Google Cloud Vision API by random Gaussian noises, suggesting random perturbation can be used in a serious adversarial attack. Meanwhile, the robustness of classifiers to various kinds of random noises, such as uniform noise in the $\ell_p$ unit ball and Gaussian noise with an arbitrary covariance matrix, has been studied in~\cite{franceschi2018robustness}. This can apply to linear classifiers as well as non-linear classifiers with locally approximately flat decision boundaries. The bounds in the uniform $\ell_p$ case depend on some universal constants, which may be arbitrarily large or small and can impact the quality of these bounds. Recently, the robustness of classifiers to perturbations under the assumption of Gaussian distributed latent input vectors has been studied in \cite{fawzi2018adversarial}. Moreover, all the results in~\cite{fawzi2018adversarial} depend on the modulus of continuity constant, which can be arbitrarily large since one cannot control it. Due to these limitations, the bounds in these recent papers cannot be directly used to deliver certified robustness metrics. We note that our probabilistic framework is not limited to supervised neural network models -- while this work is under review, a very recent workshop paper~\cite{Djverifydeepmodel} takes a similar approach to verify some properties (e.g., monotonicity and convexity) of deep probabilistic models such as variational autoencoders (VAEs)~\cite{kingma2013auto} and conditional VAEs. The key differences are that in their setting the uncertainty source is the latent variable sampled from a distribution generated by the encoder (they consider only Gaussian distribution), whereas our uncertainty comes from input perturbations (we consider Gaussian as well as general bounded distributions) and our focus is to verify neural network classifiers in supervised learning instead of generative models. This also indicates a connection between probabilistic robustness certification of neural network classifiers and property verification of deep generative models.

%% file: 2_theory.tex
\newcommand{\x}{\mathbf{x}}
\newcommand{\xo}{\mathbf{x_0}}
\newcommand{\W}[1]{\mathbf{W}^{#1}}
\newcommand{\A}[1]{\mathbf{A}^{#1}}
\newcommand{\Au}[1]{\mathbf{\Lambda}^{#1}}
\newcommand{\Al}[1]{\mathbf{\Omega}^{#1}}
\newcommand{\DD}[1]{\mathbf{D}^{#1}}
\newcommand{\Du}[1]{\mathbf{\lambda}^{#1}}
\newcommand{\Dl}[1]{\mathbf{\omega}^{#1}}
\newcommand{\Lam}[1]{\mathbf{\Lambda}^{#1}}
\newcommand{\upbias}[1]{\mathbf{\Delta}^{#1}}
\newcommand{\lwbias}[1]{\mathbf{\Theta}^{#1}}
\newcommand{\upbnd}[1]{\mathbf{u}^{#1}}
\newcommand{\lwbnd}[1]{\mathbf{l}^{#1}}
\newcommand{\z}{\mathbf{z}}
\newcommand{\y}{\mathbf{y}}
\newcommand{\bias}[1]{\mathbf{b}^{#1}}
\newcommand{\setA}{\mathcal{A}}
\newcommand{\setIpos}[1]{\mathcal{S}^{+}_{#1}}
\newcommand{\setIneg}[1]{\mathcal{S}^{-}_{#1}}
\newcommand{\setIuns}[1]{\mathcal{S}^{\pm}_{#1}}
\newcommand{\set}[1]{\mathcal{#1}}
\newcommand{\Lipsloc}{L_{q,x_0}^j}
\newcommand{\R}{\mathbb{R}}
\newcommand{\Ball}{\mathbb{B}_{p}(\xo,\epsilon)}
\newcommand{\Ph}[1]{\Phi_{#1}}
\newcommand{\upslp}[2]{\mathbf{\alpha}^{#1}_{U,{#2}}}
\newcommand{\lwslp}[2]{\mathbf{\alpha}^{#1}_{L,{#2}}}
\newcommand{\upicp}[2]{\mathbf{\beta}^{#1}_{U,{#2}}}
\newcommand{\lwicp}[2]{\mathbf{\beta}^{#1}_{L,{#2}}}

\newcommand{\Prob}[1]{\mathbb{P}\left[{#1}\right]}
\newcommand{\Qed}{$\flushright \square$}

%%% for stating theorems/lemmas/claims\Dl
\newtheorem{theorem}{Theorem}[section]
\newtheorem{lemma}[theorem]{Lemma}
\newtheorem{definition}[theorem]{Definition}
\newtheorem{notation}[theorem]{Notation}
\newtheorem{proof}[theorem]{Proof}
\newtheorem{proposition}[theorem]{Proposition}
\newtheorem{corollary}[theorem]{Corollary}
\newtheorem{conjecture}[theorem]{Conjecture}
\newtheorem{assumption}[theorem]{Assumption}
\newtheorem{observation}[theorem]{Observation}
\newtheorem{fact}[theorem]{Fact}
\newtheorem{remark}[theorem]{Remark}
\newtheorem{claim}[theorem]{Claim}
\newtheorem{example}[theorem]{Example}
\newtheorem{problem}[theorem]{Problem}
\newtheorem{open}[theorem]{Open Problem}
\newtheorem{property}[theorem]{Property}
\newtheorem{hypothesis}[theorem]{Hypothesis}

\section{PROVEN: a probabilistic framework to certify neural network robustness}
%\paragraph{Overview of our results.}
In this section, we present a general probabilistic framework \proven together with related theoretical results to compute the certified bounds in  probability that a classifier can never be fooled when the inputs of the classifier are perturbed with some given distributions. We first introduce a worst-case setting, where an input example can be perturbed by any perturbation bounded within an $\ell_p$ ball, and present corresponding worst-case analysis results.
%in Section~\ref{sec:worst-case}.
We then show that it is possible to extend these worst-case analysis results to a probabilistic setting where the input perturbations follow some given distributions, and present our probabilistic framework and main theorem.
%-- the key ideas and core theorem of our probabilistic framework is presented in Section~\ref{sec:prob_framework}. In Section~\ref{sec:evaluate},
Lastly, we provide \textit{closed-form} probabilistic bounds for various probabilistic distributions that the input perturbations can follow.

\subsection{Worst-case setting}
\label{sec:worst-case}
Let $f(\x): \R^{n_0} \rightarrow \R^{K}$ denote a $K$-class neural network classifier of interest, which takes an input $\x$ (e.g., an image) and outputs the corresponding logit scores over all classes. The ultimate goal is to efficiently find the largest $\epsilon^*$ such that the original predicted class $c$ always has a larger score $f_c(\x)$ than the score $f_t(\x)$ for a targeted attack class $t$ when the input is perturbed within the $\ell_p$ ball having radius $\epsilon^*$. Letting $g_t(\x) = f_c(\x) - f_t(\x)$, we want to find the largest $\epsilon^*$ such that $g_t(\x) > 0$ for all $\x$ satisfying $\|\x - \xo \|_p \leq  \epsilon^*$ and $c = \text{argmax}_i f_i(\xo), t \neq c$. This $\epsilon^*$ is a \textit{certified lower bound} of the minimum adversarial distortion as first introduced in Section \ref{sec:intro}.

It has been shown in \cite{weng2018towards,zhang2018efficient,Boopathy2019cnncert} that the output $f_i(\x)$ and the margin function $g_t(\x)$ of a general (convolutional) neural network classifier with general activation functions (including but not limited to ReLU, tanh, arctan, sigmoid) can be bounded by two linear functions. In other words, the authors show that  
\begin{equation}
\label{eq:g_t_relations}
    g_t^L(\x) \leq g_t(\x) \leq g_t^U(\x),
\end{equation}
where $g_t^L(\x): \R^{n_0} \rightarrow \R$ and $g_t^U(\x): \R^{n_0} \rightarrow \R$ are two linear functions
\begin{equation}
\label{eq:g_t_LU}
    g_t^L(\x) = \A{L}_{t,:} \x + d^L \quad \text{and} \quad g_t^U(\x) = \A{U}_{t,:} \x + d^U
\end{equation}
with $\A{L}_{t,:},\A{U}_{t,:} \in \R^{1\times n_0}$ being two constant row vectors and $d^L, d^U \in \R$ being two constants related to the network weights $\W{(k)}$ and biases $\bias{(k)}$ as well as the parameters bounding the activation functions in each neuron. The superscripts $L$ and $U$ denote the parameters corresponding to the lower bound and the upper bound of $g_t(\x)$.

%Since the last layer of the network is linear, the difference of the outputs $g_t(\x) = f_c(\x) - f_t(\x),\, g_t(\x): \R^{n_0} \rightarrow \R$ is also bounded by two linear functions $g_t^L(\x): \R^{n_0} \rightarrow \R$ and $g_t^U(\x): \R^{n_0} \rightarrow \R$ as follows

As the network output is bounded, the positiveness of the lower bound of $g_t(\x)$ implies that $g_t(\x)$ is positive, i.e.,
\begin{equation}
\label{eq:fast-lin-criteria}
   g_t^L(\x) > 0 \implies g_t(\x) > 0.
\end{equation}
Here, a \textit{worst-case} analysis can be performed by minimizing the linear function $g_t^L(\x)$ over all possible inputs in the set $\{\x:\|\x - \xo \|_p \leq  \epsilon \}$, which yields a closed-form solution as presented in \cite{weng2018towards,zhang2018efficient,Boopathy2019cnncert}. Therefore, the condition of whether $g_t^L(\x) > 0$ can be conveniently checked given some $\epsilon$ using the closed-form solutions; the largest $\epsilon$ such that $g_t^L(\x) > 0$ is called the \textit{certified lower bound}, which can be computed by bisection with respect to $\epsilon$.   

\subsection{Our proposed probabilistic framework: \proven}
\label{sec:prob_framework}
In addition to considering the worst-case condition for $g_t(\x) > 0$ over the norm ball constrained on the input $\{\x:\|\x - \xo \|_p \leq  \epsilon \}$, we now show that it is possible to formulate a probabilistic setting and derive bounds with guarantees by building upon the above results that the neural network output can be bounded by two linear functions. %\cite{weng2018towards,zhang2018efficient,Boopathy2019cnncert}.
We start by presenting the problem formulation of the probabilistic setting and then present our main theoretical results in Theorem~\ref{thm:prob_bnd}. 

\textbf{Problem formulation.} Consider a neural network classifier $f(\x)$ and an input example $\xo$. Let the predicted class of $\xo$ be $c$, the targeted attack class $t$, and the margin function $g_t(\x) = f_c(\x) - f_t(\x)$. Suppose the perturbed input random vector $X$ follows some given distribution $\mathcal{D}$, i.e., $X \sim \mathcal{D}$. We are interested in the probability of the margin function $g_t(\x)$ being greater than some value $a \in \R$, i.e., $\Prob{g_t(X) > a}$. 

Given that the neural network $f(\x)$ is highly non-linear and non-convex in $\x$, it is hard to directly compute the distribution of $g_t(X)$ given the input $X \sim \mathcal{D}$. Fortunately, we can still derive \textit{analytic lower and upper bounds} for $\Prob{g_t(X) > 0}$ with guarantees based on the result in the worst-case analysis that the margin function $g_t(x)$ can be bounded by two linear functions as shown in \eqref{eq:g_t_relations}. The following theorem provides such theoretical guarantees on $\Prob{g_t(X) > 0}$.
%and the proof is given in the Appendix. 

\begin{theorem}[Probabilistic bounds of network output]
\label{thm:prob_bnd}
Let $f(\x)$ be a $K$-class neural network classifier function, $\xo$ an input example, and $\epsilon$ such that $\| \x - \xo \|_p \leq \epsilon$, $p \geq 1$. Let $c = \text{argmax}_i f_i(\xo)$, let $t  (\neq c)$ be some targeted class, and define the margin function $g_t(\x) = f_c(\x) - f_t(\x)$. Suppose the input random vector $X \in \R^{n_0}$ follows some given distribution $\mathcal{D}$ with mean $\xo$ and let $a \in \R$ be some real number. There exists an explicit lower bound $\gamma_L$ and an explicit upper bound $\gamma_U$ on the probability $\Prob{g_t(X) > a}$ such that
\begin{equation}
\label{eq:thm_prob_relation}
    \gamma_L \leq \Prob{g_t(X) > a} \leq \gamma_U,
\end{equation}
where 
\begin{equation}
    \gamma_L = 1-F_{g_t^L(X)}(a), \quad \gamma_U = 1-F_{g_t^U(X)}(a),
\end{equation}
$F_Z(z)$ is the cumulative distribution function (CDF) of the random variable $Z$, and $g_t^L(\x)$, $g_t^U(\x)$ satisfy Equation \eqref{eq:g_t_relations}. 
\end{theorem}
\textit{Proof.} Let $h_1: \mathbb{R}^d \to \mathbb{R}$, $h_2: \mathbb{R}^d \to \mathbb{R}$, and $h_1(\x) \geq h_2(\x)$, $\forall \x \in \mathbb{R}^d$. Let $X \in \mathbb{R}^d$ be a random vector and let $Y_1 \in \R, Y_2 \in \R^+$ be two random variables. Since $h_1(\x) \geq h_2(\x)$ always hold, we can let $h_1(X) = Y_1 + Y_2$ and $h_2(X) = Y_1$ for some random value $Y_2 \geq 0$. We therefore have 
\begin{align*}
    \mathbb{P}[h_1(X) > a] &= \mathbb{P}[Y_1 + Y_2 > a] = \mathbb{P}[Y_1 > a - Y_2], \\
    \mathbb{P}[h_2(X) > a] &= \mathbb{P}[Y_1 > a].
\end{align*}
Since $a - Y_2 \leq a$ for any random value $Y_2 \geq 0$, we obtain $\mathbb{P}[Y_1 > a - Y_2] \geq \mathbb{P}[Y_1 > a]$ based on the fact that the cumulative distribution function is nondecreasing \cite{ShakedShanthikumar2007}, which is equivalent to
\begin{align}
\label{eq:h1h2}
    \mathbb{P}[h_1(X) > a] \geq \mathbb{P}[h_2(X) > a]. 
\end{align}
From the results in \cite{weng2018towards}, we know that the relationships in Equation \eqref{eq:g_t_relations}, i.e.,
\begin{equation*}
    g_t^L(\x) \leq g_t(\x) \leq g_t^U(\x),
\end{equation*}
satisfy $\|\x - \xo\|_p \leq \epsilon$ for all $\x$. Hence, upon applying Equation \eqref{eq:h1h2} to Equation \eqref{eq:g_t_relations} and using the fact that $\Prob{Z > a} = 1-F_Z(a)$, we obtain
\begin{equation*}
    \gamma_L \leq \Prob{g_t(X) > a} \leq \gamma_U,
\end{equation*}
with $\gamma_L = 1-F_{g_t^L(X)}(a), \; \gamma_U = 1-F_{g_t^U(X)}(a)$. 
\hfill $\square$

As discussed in Section \ref{sec:worst-case}, the neural network output and the margin function can be bounded by two linear functions \cite{weng2018towards,zhang2018efficient,Boopathy2019cnncert}. Here, we take an additional step to investigate the relationship between the margin function and its linear bounds in the probabilistic setting. Specifically, Theorem~\ref{thm:prob_bnd} shows that the probability of the neural network margin function being greater than some value $a$ can also be bounded by the CDFs of its linear bounds. Note that in the worst-case analysis of Section \ref{sec:worst-case}, we usually concern ourselves with the margin function $g_t(x) > 0$, i.e., $a = 0$. Analogously, in the probabilistic setting, we concern ourselves with the probability of the margin function $g_t(x) > 0$. This is indeed the guarantee provided by Theorem~\ref{thm:prob_bnd}: when the input $X \sim \mathcal{D}$, the result guarantees that the probability of $g_t(X) > a$ is at least $\gamma_L$ and at most $\gamma_U$.

\subsection{Evaluating the probabilistic bounds}
\label{sec:evaluate}
Theorem~\ref{thm:prob_bnd} provides us with a theoretical lower bound $\gamma_L$ and upper bound $\gamma_U$ for $\Prob{g_t(X) > a}$. In practice, we would like to numerically compute such bounds. Below we show it is possible to obtain explicit forms for $\gamma_L$ and $\gamma_U$ in terms of $\A{L}_{t,;}, \A{U}_{t,;}, d^L, d^U$, as well as the parameters of the probability distributions of input perturbations. By Theorem~\ref{thm:prob_bnd}, $\gamma_L$ and $\gamma_U$ only depend on the CDFs $F_{g_t^L(X)}$ and $F_{g_t^U(X)}$, and we observe that $g_t^L(X)$ and $g_t^U(X)$ are both linear functions of $X$ as follows:
\begin{align*}
    g_t^L(X) &= \sum_{i=1}^{n_0} \A{L}_{t,i} X_i + d^L, \\
    g_t^U(X) &= \sum_{i=1}^{n_0} \A{U}_{t,i} X_i + d^U.
\end{align*}
Hence, the problem of computing the CDFs $F_{g_t^L(X)}$ and $F_{g_t^U(X)}$ becomes a problem of computing the CDFs of a weighted sum of $X_i$ given $X \sim \mathcal{D}$. We primarily consider the following two cases:
\begin{itemize}
    \item[(i)] When $X_i$ are independent random variables with probability density function (pdf) $f_{X_i}$;
    \item[(ii)] When $X$ follows a multivariate normal distribution with mean $\xo$ and covariance $\Sigma$.
\end{itemize}
It also appears that these results may be extended to address some forms of negative correlation \cite{PanconesiSrinivasan1997,DubhashiRanjan1998}.

\subsubsection{Case (i)}
When $X_i$ are independent random variables with probability density function $f_{X_i}$, there are two approaches for computing the CDFs of the weighted sum.
\paragraph{Approach 1: Direct convolutions.} The pdf of the weighted sum is simply the convolution of the pdfs for each of the weighted random variables $\A{L}_{t,i} X_i$. Specifically, we have
\begin{equation*}
    f_{\A{L}_{t,:} X} = \Conv_{i=1}^{n_0} f_{\A{L}_{t,i} X_i},
\end{equation*}
where $\Conv_{i=1}^{N} h_i$ denotes convolution over the $N$ functions $h_1$ to $h_N$. The CDF of $\A{L}_{t,:} X$ can therefore be obtained from the pdf $f_{\A{L}_{t,:} X}$ and we obtain $F_{g^L_t(X)}(z) = F_{\A{L}_{t,:} X}(z-d^L)$; similarly, $F_{g^U_t(X)}(z) = F_{\A{U}_{t,:} X}(z-d^U)$. Hence, we have
\begin{equation*}
    \gamma_L = 1-F_{\A{L}_{t,:} X}(a-d^L), \quad \gamma_U = 1-F_{\A{U}_{t,:} X}(a-d^U).
\end{equation*}

\paragraph{Approach 2: Probabilistic inequalities.} Approach 1 is useful in cases where $n_0$ is not large. However, for large $n_0$, it might not be easy to directly compute the CDF through convolutions. For such cases, an alternative approach can be based on applying the probabilistic inequalities on the CDFs. Since we want to provide \textit{guarantees} on the desired probability in \eqref{eq:thm_prob_relation}, we need to find a lower bound on $\gamma_L$ and an upper bound on $\gamma_U$ via the probabilistic inequalities. These results are given in the following corollary.
\begin{corollary}
\label{corr:prob_ineq}
Let $X_i$ be bounded independent random variables with $X_i \in [\xo_i - \epsilon, \xo_i + \epsilon], \forall i \in [n_0]$, and symmetric around the mean $\xo_{i}$. Define 
\begin{equation*}
    \mu_L = \A{L}_{t,:} \xo + d^L, \; \mu_U = \A{U}_{t,:} \xo + d^U.
\end{equation*}
Then, we have  
\begin{align*}
    &\gamma_L 
    \geq 
    \begin{cases}
        1- \exp\left(-{\frac{(\mu_L-a)^2}{2\epsilon^2 \|\A{L}_{t,:}\|_2^2}}\right), &\text{ if } \mu_L-a \geq 0 \\
        0, &\text{ otherwise};
    \end{cases} 
    \\
    &\gamma_U
    \leq 
    \begin{cases}
        \exp\left(-{\frac{(\mu_U-a)^2}{2\epsilon^2 \|\A{U}_{t,:}\|_2^2}}\right), &\text{ if } -\mu_U+a \geq 0 \\
        1, &\text{ otherwise}.  
    \end{cases}
\end{align*}

\end{corollary}
\textit{Proof.} Let $W_i = \A{L}_{t,i} (X_i - \xo_i)$ and $\mu_L = \A{L}_{t,:} \xo + d^L$. We then have $-|\A{L}_{t,i}|\epsilon \leq W_i \leq |\A{L}_{t,i}|\epsilon$ where $W_i$ is symmetric with respect to zero since $X_i$ is symmetric. By using the fact that the sum of independent symmetric random variables is still a symmetric random variable \cite{ChowTeicher2003}, we derive
\begin{align*}
    \gamma_L &= \Prob{g_t^L(X) > a} 
    \\
    &= \Prob{\sum_{i=1}^{n_0} W_i > a - \mu_L} \\
    &= \Prob{\sum_{i=1}^{n_0} W_i < -a + \mu_L} \\
    &= 1 - \Prob{\sum_{i=1}^{n_0} W_i \geq -a + \mu_L} .
\end{align*}
From the Hoeffding inequality \cite{Resnick2014}, we obtain the following upper bound on the term $\Prob{\sum_{i=1}^{n_0} W_i \geq -a + \mu_L}$ when $-a + \mu_L>0$:
\begin{equation*}
    \Prob{\sum_{i=1}^{n_0} W_i \geq -a + \mu_L} \leq \exp\left(-{\frac{(\mu_L-a)^2}{2\epsilon^2 \|\A{L}_{t,:}\|_2^2}}\right),
\end{equation*}
and thus $\gamma_L \geq 1- \exp\left(-{\frac{(\mu_L-a)^2}{2\epsilon^2 \|\A{L}_{t,:}\|_2^2}}\right)$. When $-a + \mu_L \leq 0$, we use the trivial bound of $\gamma_L =0$. Similarly, for $\gamma_U$, we can define $\mu_U$ correspondingly and directly apply the Hoeffding inequality to obtain $\gamma_U \leq \exp\left(-{\frac{(\mu_U-a)^2}{2\epsilon^2 \|\A{U}_{t,:}\|_2^2}}\right)$, or use the trivial bound of $\gamma_U =1$.
\hfill $\square$

\subsubsection{Case (ii) } 
When $X$ follows a multivariate normal distribution with mean $\xo$ and covariance $\Sigma$, we are able to obtain an explicit form for the CDFs $F_{g_t^L(X)}$ and $F_{g_t^U(X)}$ based on the fact that the sum of normally distributed random variables still follows the normal distribution \cite{ChowTeicher2003}. Note that we include here both cases where (a) $X_i$ are independent Gaussian random variables ($\Sigma$ is a diagonal matrix) and (b) $X_i$ are correlated random variables ($\Sigma$ is a general covariance matrix and positive semidefinite). Our result is stated in the following corollary.

\begin{corollary}
\label{corr:prob_gaussian}
Let $X$ follow a multivariate normal distribution with mean $\xo$ and covariance $\Sigma$. Define 
\begin{align*}
    & \mu_L = \A{L}_{t,:} \xo + d^L, \; \sigma_L^2 = \A{L}_{t,:} \Sigma (\A{L}_{t,:})^\top ,\\
    & \mu_U = \A{U}_{t,:} \xo + d^U, \; \sigma_U^2 = \A{U}_{t,:} \Sigma (\A{U}_{t,:})^\top ,
\end{align*}
where $\top$ denotes the transpose operator. We then have 
    \begin{equation*}
        \gamma_L \approx \frac{1}{2}-\frac{1}{2} \text{erf}(\frac{a-\mu_L}{\sigma_L\sqrt{2}}), \quad
        \gamma_U \approx \frac{1}{2}-\frac{1}{2} \text{erf}(\frac{a-\mu_U}{\sigma_U\sqrt{2}})
    \end{equation*}
with $\text{erf}(\cdot)$ as the error function.
\end{corollary}

\textit{Proof.} The result is obtained in a straightforward manner from the fact \cite{ChowTeicher2003} that if $X \sim \mathcal{N}(\mu,\Sigma)$, then its linear combination $Z = w X + v$ also follows the normal distribution:  $Z \sim \mathcal{N}(w \mu + v, w \Sigma w^\top)$. The CDF of $Z$ is then given by $\frac{1}{2}(1+\text{erf}(\frac{z-\mu_Z}{\sigma_Z \sqrt{2}}))$, leading to the stated approximations. 
\hfill $\square$

\begin{remark}
Note that in our framework, all possible inputs have to lie in the $\ell_p$ ball with given radius $\epsilon$. Thus, in order to apply the Gaussian perturbation in our setting, we need to set an upper limit on the variance of the input such that $99.7 \%$ of the density is within the $\ell_p$ ball, i.e., the 3-$\sigma$ rule. See Section 4 Methods for more details.     
\end{remark}

\paragraph{Connection to $\ell_1$ and $\ell_2$ norms.} Our foregoing probabilistic analysis is established under the $\ell_\infty$ norm constraint. We note that this presented analysis can be easily extended to $\ell_1$ and $\ell_2$ norms by using the norm inequalities: $\|\x \|_1  \leq \sqrt{n_0} \| \x \|_2 \leq n_0 \| \x \|_\infty$.

% \textcolor{blue}{Lam: The following norm inequalities should be correct: $\| \x \|_\infty \leq \| \x \|_2 \leq \| \x \|_1 \leq \sqrt{n_0} \| \x \|_2 \leq n_0 \| \x \|_\infty$}

%% file: 3_simu.tex
\section{Experiments}
\begin{figure*}[h!]
     \caption{We plot the improvement of the largest $\epsilon$ certified by \proven with various confidence ($\gamma_L = \{99.99, 75, 50, 25, 5\}\%$) over the largest $\epsilon$ certified by worst-case robustness certification algorithms~\cite{weng2018towards,zhang2018efficient}. We consider both input perturbations being independent/correlated Gaussian random variables as in Case (ii) and indedepent random variables as in Case (i). The $x$-axis label in the figure: $\gamma_L$; $y$-axis label: Certification improvement of \proven over $\epsilon_{\text{worst-case}}$. The models are 2-4 layers MNIST networks with 1024 nodes per layer and ReLU actiavations.}\label{MNIST_plot_performance}
    \centering
    \begin{subfigure}[b]{0.45\textwidth}
        \includegraphics[width=\textwidth]{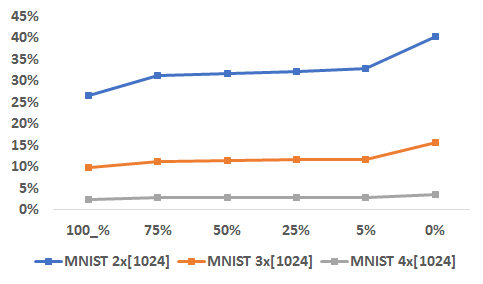}
        \caption{Case (ii) Gaussian i.i.d.}
        \label{case_01}
    \end{subfigure}
    %~ %add desired spacing between images, e. g. ~, \quad, \qquad, \hfill etc. 
    \quad
     \vspace{0.5cm}
    \begin{subfigure}[b]{0.45\textwidth}
        \includegraphics[width=\textwidth]{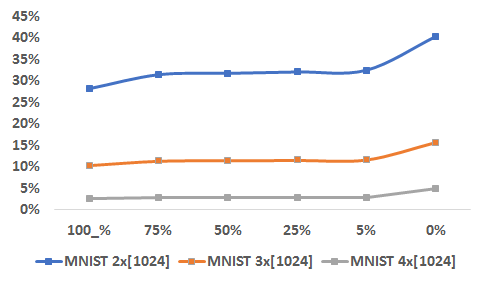}
        \caption{Case (ii) Positive correlated Gaussian}
        \label{case_02}
    \end{subfigure}
    \begin{subfigure}[b]{0.45\textwidth}
        \includegraphics[width=\textwidth]{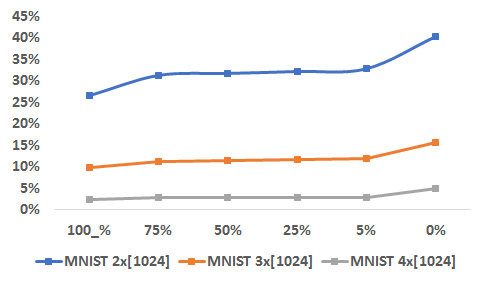}
        \caption{Case (ii) General correlated Gaussian}
        \label{case_03}
    \end{subfigure}
    \quad
    \begin{subfigure}[b]{0.45\textwidth}
        \includegraphics[width=\textwidth]{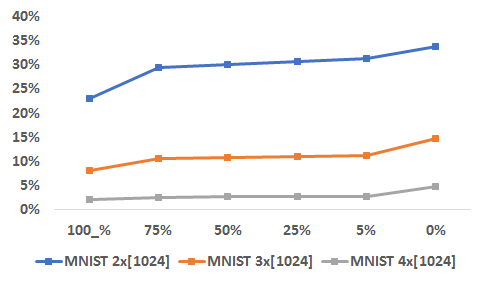}
        \caption{Case (i) Bounded independent inputs}
        \label{case_04}
    \end{subfigure}
\end{figure*}

\textbf{Methods.} We apply Corollaries~\ref{corr:prob_ineq} and~\ref{corr:prob_gaussian} to compute the largest $\epsilon$ (denoted as $\epsilon_{\proven}$) that \proven can certify with confidence of at least $\gamma_L$ when the input follows the two cases discussed in Section~\ref{sec:evaluate}. The certified lower bound computed by the worst-case analysis in \cite{weng2018towards} and~\cite{zhang2018efficient} is denoted as $\epsilon_{\text{worst-case}}$. Below is the setting of the input distributions in our simulations:
\begin{itemize}
    \item \textbf{Case (i)}. $X_i$ are independent random variables bounded in $[\xo_i-\epsilon_{\text{worst-case}},\xo_i+\epsilon_{\text{worst-case}}]$
    with mean $\xo_i$. The results are presented in Table~\ref{tab:bounded_perturbation}. 
    \item \textbf{Case (ii)}. $X$ follows a multivariate normal distribution with mean $\xo$ and covariance $\Sigma$. We consider both situations where $\Sigma$ is a positive diagonal matrix or a positive semidefinite matrix with diagonals whose square roots are less than or equal to $\epsilon_{\text{worst-case}}/3$. The results are presented in Figure~\ref{MNIST_plot_performance}.  
\end{itemize}
Note that in all the Tables, we express $\gamma_L$ as a percentage. We report $\epsilon_{\proven}$ for the following values: $\{(100-\eta), 75, 50, 25, 5, 0\} \%$ where $\eta = 10^{-2}$ and calculate the improvement of $\epsilon_{\proven}$ over $\epsilon_{\text{worst-case}}$ obtained by $(100-\eta) \%$ in the last column in Table 2 for Case (i). The results in Table 2 are averaged over 10 randomly selected images in the test sets. On the other hand, we also investigate how robust it is for the results in Table 2 by computing the average $\epsilon_{\proven}$ over randomly chosen $\{10,50,100\}$ images in 100 random trials. We report the mean and standard deviation in Table 4 and show that (a) the variation of using 10 sample average in Table 2 is less $\sim10\%$ and (b) the average $\epsilon_{\proven}$ and improvement has less deviations when we use 50 or 100 samples.

\textbf{Model and Dataset.} We use the publicly available pre-trained models provided in~\cite{weng2018towards} and~\cite{zhang2018efficient} as classifier models, which are fully-connected feed-forward neural networks with ReLU activation as well as general activations including tanh, sigmoid and arctan on the MNIST \cite{MNIST} and CIFAR-10 \cite{CIFAR10} datasets. We denote a network with $m$ layers and $n$ neurons per layer as $m \times [n]$ in the Tables.

\textbf{Implementation and Setup.} We implement \proven \footnote{https://github.com/lilyweng/PROVEN} in Python and perform experiments on a laptop with 8 Intel Cores i7-4700 HQ CPU at 2.40 GHz. 

\textbf{Result on small and large ReLU networks.} We perform simulations on both small 2-3 layer MNIST networks with 20 neurons per layer and large 2-7 layer MNIST and CIFAR networks with 1024 or 2048 neurons per layer; the results are summarized in Tables~\ref{subtab:relu} and~\ref{subtab:relu_adaptive}. These results show that on the small networks, \proven can certify up to $78.9 \%$  more with respect to the certified lower bound at the expense of decreasing the confidence by only $\eta = 10^{-2}$. In other words, \proven guarantees that at least $99.99\%$ of the $\epsilon$ computed (e.g., 0.04394 in MNIST 2$\times$[20], Table~\ref{subtab:relu}) is a certified lower bound as compared to 0.02722 for the $\epsilon_{\text{worst-case}}$ delivered by Fast-Lin\cite{weng2018towards}, where the improvement we obtained for this model is $61.4\%$. Tables~\ref{subtab:relu} and~\ref{subtab:relu_adaptive} are both ReLU activations and the only difference is the bounding techniques applied on the ReLU activations, where the bounding technique in Table~\ref{subtab:relu_adaptive} is adaptive and thus can certify more~\cite{zhang2018efficient}. For large networks, \proven can certify up to $76\%$, which is significant. Interestingly, when the bounding technique is better, it also helps our probabilistic bounds -- the improvement is significant, and even for the large CIFAR network with around 10,000 neurons, we can still obtain $10-15\%$ improvement. For the cases where the input perturbations are Gaussians, the results are presented in Figure~\ref{MNIST_plot_performance}. 

\textbf{Results on large networks with general activations.} We also ran experiments on various MNIST and CIFAR networks with non-ReLU activations, e.g., tanh, sigmoid and arctan. The results are summarized in Tables~\ref{subtab:tanh} to \ref{tab:arctan}. In comparison to the same architecture but with ReLU activations, the improvement of these activations are better than the non-adaptive bounding technique in general, and can achieve up to $32.8 \%$ on large networks. Note that the computational overhead of our approach compared to the worst-case analysis~\cite{weng2018towards,zhang2018efficient} is very little, as we only need to perform a few binary searches on the $\epsilon$ that will satisfy Corollary~\ref{corr:prob_ineq}.

\begin{table*}[t!]
\centering
\caption{The largest $\epsilon$ that \proven can certify with confidence of at least $\gamma_L = \{99.99, 75, 50, 25, 5\} \%$ when $X_i$ are independent random variables in Case (i).  We compare the largest $\epsilon$ that \proven can certify with $99.99 \%$ with the largest $\epsilon$ from state-of-the-art worst-case robustness certification algorithms~\cite{weng2018towards,zhang2018efficient} and show in the last column that \proven can certify more than the worst-case analysis by giving up $0.01\%$ confidence.}
\label{tab:bounded_perturbation}
\begin{subtable}{1\textwidth}
\caption{Relu activation\\}
\label{subtab:relu}
\begin{tabular}{l||c|ccccc|c}
\hline
Certification Method & Worst-case~\cite{weng2018towards}  &  \multicolumn{5}{c|}{Our probabilistic approach: \proven} & Certification  \\
Guarantees $\gamma_L$ & 100\%$^\dagger$  & 99.99\%$^\dagger$  & 75\%    & 50\%    & 25\%    & 5\%          & improvement$^\dagger$ \\
\hline
MNIST 2$\times${[}20{]}   & 0.02722               & \bf 0.04394 & 0.04782 & 0.04824 & 0.04859 & 0.04897  & \bf 61.4\%\\
MNIST 3$\times${[}20{]}   & 0.02127               & \bf 0.02694 & 0.02831 & 0.02847 & 0.02860 & 0.02874  & \bf 26.7\%\\
MNIST 2$\times${[}1024{]} & 0.02904               & \bf 0.03572 & 0.03758 & 0.03778 & 0.03796 & 0.03814  & \bf 23.0\%\\
MNIST 3$\times${[}1024{]} & 0.02082               & \bf 0.02253 & 0.02303 & 0.02309 & 0.02313 & 0.02318  & \bf 8.2 \% \\
MNIST 4$\times${[}1024{]} & 0.00796               & \bf 0.00813 & 0.00817 & 0.00818 & 0.00818 & 0.00818  & \bf 2.1 \% \\
CIFAR 5$\times${[}2048{]} & 0.00183               & \bf 0.00186 & 0.00186 & 0.00186 & 0.00186 & 0.00186  & \bf 1.6 \% \\
CIFAR 7$\times${[}1024{]} & 0.00189               & \bf 0.00192 & 0.00192 & 0.00193 & 0.00193 & 0.00193  & \bf 1.6 \% \\\hline
\end{tabular}
\vspace{0.4cm}
\end{subtable}
\begin{subtable}{1\textwidth}
\caption{Relu activation with adaptive bounds}
\label{subtab:relu_adaptive}
\begin{tabular}{l||c|ccccc|c}
\hline
Certification Method & Worst-case~\cite{zhang2018efficient}  &  \multicolumn{5}{c|}{Our probabilistic approach: \proven} & Certification  \\
Guarantees & 100\%$^\dagger$  & 99.99\%$^\dagger$  & 75\%    & 50\%    & 25\%    & 5\%         & improvement$^\dagger$ \\
\hline
MNIST 2$\times${[}20{]}   & 0.02746    & \bf 0.04912     & 0.05212    & 0.05246   & 0.05276   & 0.05307  & \bf 78.9 \%\\
MNIST 3$\times${[}20{]}   & 0.02236    & \bf 0.03828     & 0.03966    & 0.03981   & 0.03995   & 0.04009  & \bf 71.2 \% \\
MNIST 2$\times${[}1024{]} & 0.03158               & \bf 0.05560 & 0.05756 & 0.05779 & 0.05798 & 0.05818  & \bf 76.1 \%\\
MNIST 3$\times${[}1024{]} & 0.02397               & \bf 0.03524 & 0.03583 & 0.03589 & 0.03595 & 0.03601  & \bf 47.1 \% \\
MNIST 4$\times${[}1024{]} & 0.00962               & \bf 0.01288 & 0.01293 & 0.01294 & 0.01295 & 0.01295  & \bf 33.9 \% \\
CIFAR 5$\times${[}2048{]} & 0.00228               & \bf 0.00264 & 0.00265 & 0.00265 & 0.00265 & 0.00265  & \bf 15.8 \% \\
CIFAR 7$\times${[}1024{]} & 0.00189    & \bf 0.00209     & 0.00210    & 0.00210   & 0.00210   & 0.00210  & \bf 10.6 \%\\
\hline
\end{tabular}
\vspace{0.4cm}
\end{subtable}
\vspace{0.4cm}
\begin{subtable}{1\textwidth}
\caption{Tanh activation}
\label{subtab:tanh}
\begin{tabular}{l||c|ccccc|c}
\hline
Certification Method & Worst-case~\cite{zhang2018efficient}  &  \multicolumn{5}{c|}{Our probabilistic approach: \proven} & Certification  \\
Guarantees & 100\%$^\dagger$  & 99.99\%$^\dagger$  & 75\%    & 50\%    & 25\%    & 5\%        & improvement$^\dagger$ \\
\hline
% MNIST 2$\times${[}20{]}   &     &     &     &    &     &        &    & \\
% MNIST 3$\times${[}20{]}   &     &     &     &    &     &        &    & \\
MNIST 2$\times${[}1024{]} & 0.02232               & \bf 0.02915 & 0.03005 & 0.03013 & 0.03022 & 0.03033  & \bf 30.6\%\\
MNIST 3$\times${[}1024{]} & 0.01121               & \bf 0.01360 & 0.01376 & 0.01378 & 0.01380 & 0.01381  & \bf 21.3 \% \\
MNIST 4$\times${[}1024{]} & 0.00682               & \bf 0.00745 & 0.00750 & 0.00750 & 0.00751 & 0.00751  & \bf 9.2 \% \\
CIFAR 5$\times${[}2048{]}   & 0.00081     & \bf 0.00085   & 0.00085    & 0.00085    &  0.00085   & 0.00085         & \bf 4.9 \% \\
% CIFAR 7$\times${[}1024{]}   &     &     &     &    &     &        &    & \\
\hline
\end{tabular}
\end{subtable}
\vspace{0.4cm}
\begin{subtable}{1\textwidth}
\caption{Sigmoid activation}
\label{subtab:sigmoid}
\begin{tabular}{l||c|ccccc|c}
\hline
Certification Method & Worst-case~\cite{zhang2018efficient}  &  \multicolumn{5}{c|}{Our probabilistic approach: \proven} & Certification  \\
Guarantees & 100\%$^\dagger$  & 99.99\%$^\dagger$  & 75\%    & 50\%    & 25\%    & 5\%        & improvement$^\dagger$ \\
\hline
% MNIST 2$\times${[}20{]}   &     &     &     &    &     &        &    & \\
% MNIST 3$\times${[}20{]}   &     &     &     &    &     &        &    & \\
MNIST 2$\times${[}1024{]} & 0.02785               & \bf 0.03285 & 0.03404 & 0.03419 & 0.03426 & 0.03441 & \bf 18.0\%\\
MNIST 3$\times${[}1024{]} & 0.01856               & \bf 0.02296 & 0.02342 & 0.02348 & 0.02353 & 0.02358 & \bf 23.7 \% \\
MNIST 4$\times${[}1024{]} & 0.01778               & \bf 0.02170 & 0.02224 & 0.02229 & 0.02232 & 0.02237 & \bf 22.1 \% \\
% CIFAR 7$\times${[}1024{]}   &     &     &     &    &     &        &    & \\
\hline
\end{tabular}
\end{subtable}
\begin{subtable}{1\textwidth}
\caption{Arctan activation }
\label{tab:arctan}
\begin{tabular}{l||c|ccccc|c}
\hline
Certification Method & Worst-case~\cite{zhang2018efficient}  &  \multicolumn{5}{c|}{Our probabilistic approach: \proven} & Certification  \\
Guarantees & 100\%$^\dagger$  & 99.99\%$^\dagger$  & 75\%    & 50\%    & 25\%    & 5\%         & improvement$^\dagger$ \\
\hline
% MNIST 2$\times${[}20{]}   &     &     &     &    &     &        &    & \\
% MNIST 3$\times${[}20{]}   &     &     &     &    &     &        &    & \\
MNIST 2$\times${[}1024{]} & 0.02105               & \bf 0.02796 & 0.02907 & 0.02915 & 0.02924 & 0.02936  & \bf 32.8\%\\
MNIST 3$\times${[}1024{]} & 0.01250               & \bf 0.01462 & 0.01486 & 0.01488 & 0.01490 & 0.01493  & \bf 17.0 \% \\
MNIST 4$\times${[}1024{]} & 0.00726               & \bf 0.00829 & 0.00836 & 0.00837 & 0.00838 & 0.00838  & \bf 14.2 \% \\
CIFAR 5$\times${[}2048{]}   & 0.00078    & \bf 0.00089   & 0.00089    & 0.00089   &     0.00089 &  0.00089    & \bf 14.1 \% \\
% CIFAR 7$\times${[}1024{]}   &     &     &     &    &     &        &    & \\
\hline
\end{tabular}
\end{subtable}
\end{table*}

% \begin{figure*}[!]
%  \centering
%  \includegraphics[width=0.45\textwidth]{Figs/case_01}
%  \includegraphics[width=0.45\textwidth]{Figs/case_02}
%  \includegraphics[width=0.45\textwidth]{Figs/case_03} 
%  \includegraphics[width=0.45\textwidth]{Figs/case_04} 
%   \caption{}
%   \label{xxx}
%  \end{figure*}

% ===================================================================

% \begin{table*}[t!]
% \centering
% \caption{\textcolor{red}{New New table: general activation, increase of bounds} Input is \textcolor{blue}{r.v. with mean $\xo$ and bounded perturbation within $\epsilon_{\text{worst-case}}$}.}
% \label{tab:uniform}
% \begin{tabular}{c|ccccc}
% \hline
% model        & Relu & Relu-ada   & tanh    & sigmoid    & arctan & Relu & Relu-ada   & tanh    & sigmoid    & arctan    \\
% \hline
% MNIST 2$\times${[}1024{]} & 0.02232               &  0.02915 & 0.03005 & 0.03013 & 0.03022 \\
% MNIST 3$\times${[}1024{]} & 0.01121               &  0.01360 & 0.01376 & 0.01378 & 0.01380 \\
% MNIST 4$\times${[}1024{]} & 0.00682               &  0.00745 & 0.00750 & 0.00750 & 0.00751 \\\hline
% \end{tabular}
% \end{table*}

\begin{table*}[!]
\centering
\caption{Summary of the improvement of our approach (we certify the bound with at least $99.99 \%$ confidence) compared to $\epsilon_{\text{worst-case}}$~\cite{zhang2018efficient}.}
\label{tab:uniform}
\begin{tabular}{c|ccccc}
\hline
model        & Relu & Relu-ada   & tanh    & sigmoid    & arctan    \\
\hline
MNIST 2$\times${[}1024{]} & 23.0\%              &  76.1\% & 30.6\% & 18.0\% & 32.8\% \\
MNIST 3$\times${[}1024{]} & 8.2\%             &  47.1\% & 21.3\% & 23.7\% & 17.0\% \\
MNIST 4$\times${[}1024{]} & 2.1\%               &  33.9\% & 9.2\% & 22.1\% & 14.2\% \\\hline
\end{tabular}
\end{table*}

\begin{table*}[h!]
\centering
\caption{With input perturbations being independent random variables in case (i), we randomly choose $\{10,50,100\}$ input samples (images) in each trial and then compute the average of the largest $\epsilon$ that can be certified by worst-case analysis~\cite{zhang2018efficient} (denoted as $\epsilon_{\text{worst-case}}$) and by \proven with $99.99\%$ confidence (denoted as $\epsilon_{\proven}$) together with the improved certification of $\epsilon_{\proven}$ over $\epsilon_{\text{worst-case}}$ (denoted as Improv.). We present the mean and std of the average $\epsilon$ and the improvements for $\{10,50,100\}$ samples in a total of 100 random trials, showing that the mean and std converge as the number of samples increases.} 
\label{tab:boundwith100_trials}
\begin{subtable}{1\textwidth}
\caption{MNIST 3$\times$[1024], ReLU activation with adaptive bounds}
\label{subtab:mnist_3_relu}
\begin{tabular}{l||ccc|ccc|ccc}
\hline
\multirow{2}{*}{100 rand trials} & \multicolumn{3}{c|}{10 samples}  &  \multicolumn{3}{c|}{50 samples} & \multicolumn{3}{c}{100 samples} \\
& $\epsilon_{\text{worst-case}}$  & $\epsilon_{\proven}$  & Improv.    & $\epsilon_{\text{worst-case}}$    & $\epsilon_{\proven}$    & Improv.     & $\epsilon_{\text{worst-case}}$     & $\epsilon_{\proven}$ & Improv. \\
\hline
Mean & 0.02559  & 0.03703 & 44.75\% & 0.02581 & 0.03734 & 44.70\% & 0.02579 & 0.03733 & 44.74\% \\
std  & 0.00165  & 0.00222 & 1.12\%  & 0.00076 & 0.00102 & 0.57\%  & 0.00054 & 0.00071 & 0.43\% \\
\hline
\end{tabular}
\vspace{0.4cm}
\end{subtable}
\vspace{0.4cm}
\begin{subtable}{1\textwidth}
\caption{MNIST 3$\times$[1024], tanh activation}
\label{subtab:mnist_3_tanh}
\begin{tabular}{l||ccc|ccc|ccc}
\hline
\multirow{2}{*}{100 rand trials} & \multicolumn{3}{c|}{10 samples}  &  \multicolumn{3}{c|}{50 samples} & \multicolumn{3}{c}{100 samples} \\
& $\epsilon_{\text{worst-case}}$  & $\epsilon_{\proven}$  & Improv.    & $\epsilon_{\text{worst-case}}$    & $\epsilon_{\proven}$    & Improv.     & $\epsilon_{\text{worst-case}}$     & $\epsilon_{\proven}$ & Improv. \\
\hline
Mean & 0.01195  & 0.01375 & 15.17\% & 0.01193 & 0.01374 & 15.22\% & 0.01192 & 0.01374 & 15.25\% \\
std  & 0.00065  & 0.00068 & 2.66\%  & 0.00030 & 0.00030 & 1.27\%  & 0.00020 & 0.00021 & 0.77\% \\
\hline
\end{tabular}
\vspace{0.4cm}
\end{subtable}
\vspace{0.4cm}
\begin{subtable}{1\textwidth}
\caption{MNIST 4$\times$[1024], ReLU activation with adaptive bounds}
\label{subtab:mnist_4_relu}
\begin{tabular}{l||ccc|ccc|ccc}
\hline
\multirow{2}{*}{100 rand trials} & \multicolumn{3}{c|}{10 samples}  &  \multicolumn{3}{c|}{50 samples} & \multicolumn{3}{c}{100 samples} \\
& $\epsilon_{\text{worst-case}}$  & $\epsilon_{\proven}$  & Improv.    & $\epsilon_{\text{worst-case}}$    & $\epsilon_{\proven}$    & Improv.     & $\epsilon_{\text{worst-case}}$     & $\epsilon_{\proven}$ & Improv. \\
\hline
Mean & 0.00998  & 0.01329 & 33.18\% & 0.00994 & 0.01325 & 33.24\% & 0.00997 & 0.01328 & 33.21\% \\
std  & 0.00051  & 0.00066 & 0.57\%  & 0.00021 & 0.00027 & 0.27\%  & 0.00014 & 0.00018 & 0.15\% \\
\hline
\end{tabular}
\vspace{0.4cm}
\end{subtable}
\vspace{0.4cm}
\begin{subtable}{1\textwidth}
\caption{MNIST 3$\times$[1024], tanh activation}
\label{subtab:cifar_5_tanh}
\begin{tabular}{l||ccc|ccc|ccc}
\hline
\multirow{2}{*}{100 rand trials} & \multicolumn{3}{c|}{10 samples}  &  \multicolumn{3}{c|}{50 samples} & \multicolumn{3}{c}{100 samples} \\
& $\epsilon_{\text{worst-case}}$  & $\epsilon_{\proven}$  & Improv.    & $\epsilon_{\text{worst-case}}$    & $\epsilon_{\proven}$    & Improv.     & $\epsilon_{\text{worst-case}}$     & $\epsilon_{\proven}$ & Improv. \\
\hline
Mean & 0.01195  & 0.01375 & 15.17\% & 0.01193 & 0.01374 & 15.22\% & 0.01192 & 0.01374 & 15.25\% \\
std  & 0.00065  & 0.00068 & 2.66\%  & 0.00030 & 0.00030 & 1.27\%  & 0.00020 & 0.00021 & 0.77\% \\
\hline
\end{tabular}
\vspace{0.4cm}
\end{subtable}
\vspace{0.4cm}
\begin{subtable}{1\textwidth}
\caption{CIFAR 5$\times$[2048], ReLU activation with adaptive bounds}
\label{subtab:cifar_5_relu}
\begin{tabular}{l||ccc|ccc|ccc}
\hline
\multirow{2}{*}{100 rand trials} & \multicolumn{3}{c|}{10 samples}  &  \multicolumn{3}{c|}{50 samples} & \multicolumn{3}{c}{100 samples} \\
& $\epsilon_{\text{worst-case}}$  & $\epsilon_{\proven}$  & Improv.    & $\epsilon_{\text{worst-case}}$    & $\epsilon_{\proven}$    & Improv.     & $\epsilon_{\text{worst-case}}$     & $\epsilon_{\proven}$ & Improv. \\
\hline
Mean & 0.00224  & 0.00264 & 18.07\% & 0.00222 & 0.00262 & 17.93\% & 0.00222 & 0.00263 & 18.06\% \\
std  & 0.00020  & 0.00025 &  2.39\% & 0.00009 & 0.00011 &  1.12\% & 0.00005 & 0.00006 &  0.55\% \\
\hline
\end{tabular}
\vspace{0.4cm}
\end{subtable}
\vspace{0.4cm}
\begin{subtable}{1\textwidth}
\caption{CIFAR 5$\times$[2048], arctan activation}
\label{subtab:cifar_5_arctan}
\begin{tabular}{l||ccc|ccc|ccc}
\hline
\multirow{2}{*}{100 rand trials} & \multicolumn{3}{c|}{10 samples}  &  \multicolumn{3}{c|}{50 samples} & \multicolumn{3}{c}{100 samples} \\
& $\epsilon_{\text{worst-case}}$  & $\epsilon_{\proven}$  & Improv.    & $\epsilon_{\text{worst-case}}$    & $\epsilon_{\proven}$    & Improv.     & $\epsilon_{\text{worst-case}}$     & $\epsilon_{\proven}$ & Improv. \\
\hline
Mean & 0.00091  & 0.00100 & 9.28\% & 0.00091 & 0.00100 & 9.32\% & 0.00092 & 0.00100 & 9.32\% \\
std  & 0.00008  & 0.00009 & 3.17\% & 0.00003 & 0.00003 & 1.15\% & 0.00001 & 0.00002 & 0.56\% \\
\hline
\end{tabular}
\vspace{0.4cm}
\end{subtable}
\vspace{0.4cm}
\begin{subtable}{1\textwidth}
\caption{CIFAR 7$\times$[1024], ReLU activation with adaptive bound}
\label{subtab:cifar_7_relu}
\begin{tabular}{l||ccc|ccc|ccc}
\hline
\multirow{2}{*}{100 rand trials} & \multicolumn{3}{c|}{10 samples}  &  \multicolumn{3}{c|}{50 samples} & \multicolumn{3}{c}{100 samples} \\
& $\epsilon_{\text{worst-case}}$  & $\epsilon_{\proven}$  & Improv.    & $\epsilon_{\text{worst-case}}$    & $\epsilon_{\proven}$    & Improv.     & $\epsilon_{\text{worst-case}}$     & $\epsilon_{\proven}$ & Improv. \\
\hline
Mean & 0.00176  & 0.00195 & 10.68\% & 0.00174 & 0.00192 & 10.73\% & 0.00174 & 0.00193 & 10.70\% \\
std  & 0.00018  & 0.00020 &  1.87\% & 0.00007 & 0.00008 &  0.75\% & 0.00003 & 0.00004 &  0.37\% \\
\hline
\end{tabular}
\vspace{0.4cm}
\end{subtable}
\end{table*}

%% file: 4_conclusion.tex
\section{Conclusions and future works}
We proposed a novel probabilistic framework \proven to certify the robustness of neural networks and derived theoretical bounds on the robustness certification with statistical guarantees. \proven is a general tool that can build on top of existing state-of-the-art neural network robustness certification algorithms (Fast-Lin, CROWN and CNN-Cert) and hence can be readily applied to certify fully-connected and convolutional neural networks with different activation functions. Experimental results on large neural networks demonstrated significant benefits of \proven over the standard worst-case analysis results.

%% file: main.bbl
\begin{thebibliography}{10}
\providecommand{\url}[1]{#1}
\csname url@samestyle\endcsname
\providecommand{\newblock}{\relax}
\providecommand{\bibinfo}[2]{#2}
\providecommand{\BIBentrySTDinterwordspacing}{\spaceskip=0pt\relax}
\providecommand{\BIBentryALTinterwordstretchfactor}{4}
\providecommand{\BIBentryALTinterwordspacing}{\spaceskip=\fontdimen2\font plus
\BIBentryALTinterwordstretchfactor\fontdimen3\font minus
  \fontdimen4\font\relax}
\providecommand{\BIBforeignlanguage}[2]{{%
\expandafter\ifx\csname l@#1\endcsname\relax
\typeout{** WARNING: IEEEtran.bst: No hyphenation pattern has been}%
\typeout{** loaded for the language `#1'. Using the pattern for}%
\typeout{** the default language instead.}%
\else
\language=\csname l@#1\endcsname
\fi
#2}}
\providecommand{\BIBdecl}{\relax}
\BIBdecl

\bibitem{szegedy2013intriguing}
C.~Szegedy, W.~Zaremba, I.~Sutskever, J.~Bruna, D.~Erhan, I.~Goodfellow, and
  R.~Fergus, ``Intriguing properties of neural networks,'' \emph{arXiv preprint
  arXiv:1312.6199}, 2013.

\bibitem{cisse2017houdini}
M.~M. Cisse, Y.~Adi, N.~Neverova, and J.~Keshet, ``Houdini: Fooling deep
  structured visual and speech recognition models with adversarial examples,''
  in \emph{NIPS}, 2017.

\bibitem{wang2017adversary}
Q.~Wang, W.~Guo, K.~Zhang, A.~G. Ororbia~II, X.~Xing, X.~Liu, and C.~L. Giles,
  ``Adversary resistant deep neural networks with an application to malware
  detection,'' in \emph{SIGKDD}.\hskip 1em plus 0.5em minus 0.4em\relax ACM,
  2017.

\bibitem{chen2018ordered}
P.-Y. Chen, B.~Vinzamuri, and S.~Liu, ``Is ordered weighted $\ell_1$
  regularized regression robust to adversarial perturbation? a case study on
  oscar,'' \emph{arXiv preprint arXiv:1809.08706}, 2018.

\bibitem{goodfellow2014explaining}
I.~J. Goodfellow, J.~Shlens, and C.~Szegedy, ``Explaining and harnessing
  adversarial examples,'' \emph{ICLR, arXiv preprint arXiv:1412.6572}, 2015.

\bibitem{biggio2017wild}
B.~Biggio and F.~Roli, ``Wild patterns: Ten years after the rise of adversarial
  machine learning,'' \emph{arXiv preprint arXiv:1712.03141}, 2017.

\bibitem{katz2017reluplex}
G.~Katz, C.~Barrett, D.~L. Dill, K.~Julian, and M.~J. Kochenderfer, ``Reluplex:
  An efficient smt solver for verifying deep neural networks,'' in
  \emph{International Conference on Computer Aided Verification}.\hskip 1em
  plus 0.5em minus 0.4em\relax Springer, 2017, pp. 97--117.

\bibitem{sinha2017certifiable}
A.~Sinha, H.~Namkoong, and J.~Duchi, ``Certifiable distributional robustness
  with principled adversarial training,'' \emph{ICLR, arXiv preprint
  arXiv:1710.10571}, 2018.

\bibitem{ehlers2017formal}
R.~Ehlers, ``Formal verification of piece-wise linear feed-forward neural
  networks,'' in \emph{International Symposium on Automated Technology for
  Verification and Analysis}.\hskip 1em plus 0.5em minus 0.4em\relax Springer,
  2017, pp. 269--286.

\bibitem{hein2017formal}
M.~Hein and M.~Andriushchenko, ``Formal guarantees on the robustness of a
  classifier against adversarial manipulation,'' in \emph{Advances in Neural
  Information Processing Systems}, 2017, pp. 2263--2273.

\bibitem{weng2018evaluating}
T.-W. Weng, H.~Zhang, P.-Y. Chen, J.~Yi, D.~Su, Y.~Gao, C.-J. Hsieh, and
  L.~Daniel, ``Evaluating the robustness of neural networks: An extreme value
  theory approach,'' \emph{ICLR, arXiv preprint arXiv:1801.10578}, 2018.

\bibitem{kolter2017provable}
J.~Z. Kolter and E.~Wong, ``Provable defenses against adversarial examples via
  the convex outer adversarial polytope,'' \emph{arXiv preprint
  arXiv:1711.00851}, 2017.

\bibitem{raghunathan2018certified}
A.~Raghunathan, J.~Steinhardt, and P.~Liang, ``Certified defenses against
  adversarial examples,'' \emph{ICLR, arXiv preprint arXiv:1801.09344}, 2018.

\bibitem{weng2018towards}
T.-W. Weng, H.~Zhang, H.~Chen, Z.~Song, C.-J. Hsieh, D.~Boning, I.~S. Dhillon,
  and L.~Daniel, ``Towards fast computation of certified robustness for relu
  networks,'' \emph{ICML, arXiv preprint arXiv:1804.09699}, 2018.

\bibitem{zhang2018efficient}
H.~Zhang, T.-W. Weng, P.-Y. Chen, C.-J. Hsieh, and L.~Daniel, ``Efficient
  neural network robustness certification with general activation functions,''
  in \emph{Advances in Neural Information Processing Systems (NIPS)}, 2018.

\bibitem{Gehr2018AI2}
T.~Gehr, M.~Mirman, D.~Drachsler-Cohen, P.~Tsankov, S.~Chaudhuri, and
  M.~Vechev, ``Ai2: Safety and robustness certification of neural networks with
  abstract interpretation,'' in \emph{IEEE Symposium on Security and Privacy
  (SP)}, vol.~00, 2018, pp. 948--963.

\bibitem{Boopathy2019cnncert}
A.~Boopathy, T.-W. Weng, P.-Y. Chen, S.~Liu, and L.~Daniel, ``Cnn-cert: An
  efficient framework for certifying robustness of convolutional neural
  networks,'' in \emph{AAAI}, Jan 2019.

\bibitem{franceschi2018robustness}
J.-Y. Franceschi, A.~Fawzi, and O.~Fawzi, ``Robustness of classifiers to
  uniform $\ell_p$ and gaussian noise,'' \emph{arXiv preprint
  arXiv:1802.07971}, 2018.

\bibitem{fawzi2018adversarial}
A.~Fawzi, H.~Fawzi, and O.~Fawzi, ``Adversarial vulnerability for any
  classifier,'' \emph{arXiv preprint arXiv:1802.08686}, 2018.

\bibitem{dvijotham2018dual}
K.~Dvijotham, R.~Stanforth, S.~Gowal, T.~Mann, and P.~Kohli, ``A dual approach
  to scalable verification of deep networks,'' \emph{arXiv preprint
  arXiv:1803.06567}, 2018.

\bibitem{peck2017lower}
J.~Peck, J.~Roels, B.~Goossens, and Y.~Saeys, ``Lower bounds on the robustness
  to adversarial perturbations,'' in \emph{Advances in Neural Information
  Processing Systems (NIPS)}, 2017, pp. 804--813.

\bibitem{bibi2018analytic}
A.~Bibi, M.~Alfadly, and B.~Ghanem, ``Analytic expressions for probabilistic
  moments of pl-dnn with gaussian input,'' in \emph{The IEEE Conference on
  Computer Vision and Pattern Recognition (CVPR)}, 2018.

\bibitem{fawzi2016robustness}
A.~Fawzi, S.-M. Moosavi-Dezfooli, and P.~Frossard, ``Robustness of classifiers:
  from adversarial to random noise,'' in \emph{Advances in Neural Information
  Processing Systems}, 2016, pp. 1632--1640.

\bibitem{hosseini2017google}
H.~Hosseini, B.~Xiao, and R.~Poovendran, ``Google's cloud vision api is not
  robust to noise,'' in \emph{Machine Learning and Applications (ICMLA), 2017
  16th IEEE International Conference on}.\hskip 1em plus 0.5em minus
  0.4em\relax IEEE, 2017, pp. 101--105.

\bibitem{Djverifydeepmodel}
K.~Dvijotham, M.~Garnelo, A.~Fawzi, and P.~Kohli, ``Verification of deep
  probabilistic models,'' in \emph{arXiv preprint arXiv:1812.02795}, 2018.

\bibitem{kingma2013auto}
D.~P. Kingma and M.~Welling, ``Auto-encoding variational bayes,'' \emph{arXiv
  preprint arXiv:1312.6114}, 2013.

\bibitem{ShakedShanthikumar2007}
M.~Shaked and G.~Shanthikumar, \emph{Stochastic Orders}.\hskip 1em plus 0.5em
  minus 0.4em\relax Springer, 2007.

\bibitem{PanconesiSrinivasan1997}
A.~Panconesi and A.~Srinivasan, ``Randomized distributed edge coloring via an
  extension of the chernoff-hoeffding bounds,'' \emph{{SIAM} Journal on
  Computing}, vol.~26, no.~2, pp. 350--368, 1997.

\bibitem{DubhashiRanjan1998}
D.~P. Dubhashi and D.~Ranjan, ``Balls and bins: A study in negative
  dependence,'' \emph{Random Structures and Algorithms}, vol.~13, no.~2, pp.
  99--124, 1998.

\bibitem{ChowTeicher2003}
Y.~S. Chow and H.~Teicher, \emph{Probability Theory: Independence,
  Interchangeability, Martingales}, 3rd~ed.\hskip 1em plus 0.5em minus
  0.4em\relax Springer, 2003.

\bibitem{Resnick2014}
S.~I. Resnick, \emph{A Probability Path}.\hskip 1em plus 0.5em minus
  0.4em\relax Birkh{\"a}user, 2014.

\bibitem{MNIST}
Y.~LeCun, L.~Bottou, Y.~Bengio, and P.~Haffner, ``Gradient-based learning
  applied to document recognition,'' \emph{Proceedings of the IEEE}, vol.~86,
  no.~11, pp. 2278--2324, 1998.

\bibitem{CIFAR10}
A.~Krizhevsky and G.~Hinton, ``Learning multiple layers of features from tiny
  images,'' Citeseer, Tech. Rep., 2009.

\end{thebibliography}
